\documentstyle{llncs}
\RequirePackage{graphicx}

\def\Var{{\rm Var}}

\begin{document}
\title{Calibration with Changing Checking Rules and Its Application to
Short-Term Trading}

\author{Vladimir Trunov and Vladimir V'yugin}

\institute{Institute for Information Transmission Problems,
Russian Academy of Sciences,
Bol'shoi Karetnyi per. 19, Moscow GSP-4, 127994, Russia\\
email: vyugin@iitp.ru}

\maketitle

\begin{abstract}
We provide a natural learning process in which a financial
trader without a risk receives
a gain in case when Stock Market is inefficient.
In this process, the trader rationally choose his gambles
using a prediction made by a randomized calibrated
algorithm. Our strategy is based on Dawid's notion
of calibration with more general changing checking rules and on some
modification of Kakade and Foster's
randomized algorithm for computing calibrated forecasts.
\end{abstract}

\section{Introduction}\label{intr-1}

Predicting sequences is the key problem of machine learning and
statistics. The learning process proceeds as follows: observing a finite-state
sequence given on-line a forecaster assigns an subjective estimate to
future states. The method of evaluation of these forecasts depends on an
underlying learning approach.

A minimal requirement for testing any prediction algorithm is that
it should be calibrated (see Dawid~\cite{Daw82}).
Dawid gave an informal explanation of calibration for binary outcomes
as follows.
Let a binary sequence $\omega_1,\omega_2,\dots ,\omega_{n-1}$ of outcomes
be observed by a forecaster whose task is to give a probability $p_n$
of a future event $\omega_n=1$. In a typical example, $p_n$ is interpreted
as a probability that it will rain. Forecaster is said to be
well-calibrated if it rains as often as he leads us to expect. It should rain
about $80\%$ of the days for which $p_n=0.8$, and so on.

A more precise definition is as follows.
Let $I(p)$ denote the characteristic function of a
subinterval $I\subseteq [0,1]$, i.e., $I(p)=1$ if $p\in I$, and $I(p)=0$,
otherwise.
An infinite sequence of forecasts $p_1,p_2,\dots$ is calibrated
for an infinite binary sequence of outcomes
$\omega_1\omega_2\dots$ if for characteristic function $I(p)$
of any subinterval of $[0,1]$ the calibration error tends to zero, i.e.,
\begin{eqnarray*}\label{call-1}
\frac{\sum_{i=1}^{n}I(p_i)(\omega_i-p_i)}
{\sum_{i=1}^{n}I(p_i)}\to 0
\end{eqnarray*}
as the denominator of the relation (\ref{call-1}) tends to infinity.

The indicator function $I(p_i)$ determines some
``checking rule'' which selects indices $i$ where we compute
the deviation between forecasts $p_i$ and outcomes $\omega_i$.

If the weather acts adversatively, then Oakes~\cite{Oak85} and
Dawid~\cite{Daw85} show that any deterministic forecasting
algorithm will not always be calibrated.

Foster and Vohra~\cite{FoV98} show that calibration is almost surely
guaranteed with a randomizing forecasting rule, i.e., where the forecasts
$p_i$ are chosen using internal randomization and the forecasts are hidden
from the weather until weather makes its decision whether to rain or not.

The origin of calibration algorithm is the Blackwell's~\cite{Bla56}
approachability theorem
but, as its drawback, the forecaster has to use linear programming to compute
the forecasts. We modify a more computationally efficient method from
Kakade and Foster~\cite{KaF2004}, where ``an almost deterministic''
randomized rounding universal forecasting algorithm is presented.
For any sequence of outcomes and for any precision of rounding $\Delta>0$,
an observer can simply randomly round the deterministic forecast $p_i$ up to
$\Delta$ in order to calibrate for this sequence with probability one~:
\begin{eqnarray}\label{call-1a}
\limsup\limits_{n\to\infty}\frac{1}{n}\sum_{i=1}^{n}
I(\tilde p_i)(\omega_i-\tilde p_i)\le\Delta,
\end{eqnarray}
where $\tilde p_i$ is a random forecast.
This algorithm can be easily extended
such that the calibration error tends to zero as $n\to\infty$.

The goal of this paper is to extend Kakade and Foster's algorithm
to arbitrary real valued outcomes and to a more general notion of calibration
with changing parameterized checking rules.
We present also convergence bounds for
calibration depending on the number of parameters.

A closely related approach for weak calibration is presented
in Vovk~\cite{Vov2007}.

We apply this algorithm to technical analysis in finance.
We consider real valued outcomes (for example, prices of a stock).
In this case,
predictions could be interpreted as mean values of future outcomes under
some unknown for us probability distributions. We need not
any form of such distribution -- we should predict only future means.

We provide a natural learning process in which a financial
trader (speculator for a rise or for a decline) without a risk of
complete ruin receives a gain if the market is inefficient.
In this process, the trader rationally choose his gambles
using a prediction made by a randomized calibrated algorithm.

The learning process is the most traditional one. At each step
{\it Forecaster} makes a prediction of future price of a stock
and {\it Speculator} takes the best response to this
prediction. He chooses a strategy: dealing for a rise or for a
fall, or pass the step. {\it Forecaster} uses some randomized
algorithm for computing calibrated forecasts.

Let us give a more precise formulation.
Consider a game between {\it Speculator} and {\it Stock Market}.
Let $S_1,S_2,\dots$ -- be a sequence of prices of a stock.
We suppose that prices are bounded and rescaled such that
$0\le S_i\le 1$ for all $t$ and $S_1=S_0$.

The protocol of a game is described as follows. Let $k$ is a positive
integer number.
The initial capital of {\it Speculator} is ${\cal K}_0=0$.

\noindent{\bf FOR} $i=1,2\dots$
\\
At the beginning of the step $i$ {\it Speculator} and {\it Forecaster}
observe past prices $S_1,\dots ,S_{i-1}$ of a financial instrument (a stock)
and some side information.
\\
{\it Forecaster} announces a random forecast of a stock
future price -- random variable $\tilde p_i\in [0,1]$.
\\
{\it Speculator} bets by buying or selling a number $M_i$ of shares of
the stock by $S_{i-1}$ each.
\footnote{In case $M_i>0$ {\it Speculator} playing for a rise, in case
$M_i<0$ {\it Speculator} playing for a fall, {\it Speculator} pass the step
if $M_i=0$.
We suppose that {\it Speculator} can borrow money for
buying $M_i$ shares of a stock and return them after selling.}
\\
{\it Stock Market} announces a price $S_i$ of a stock.
\\
{\it Speculator} receives his total gain (or suffer loss)
at the end of step $i$~~:
\\
${\cal K}_i={\cal K}_{i-1}+M_i (S_i-S_{i-1})$.
\\
{\bf ENDFOR}

In that follows we consider only playing for a rise and
will buy only one share of a stock, so $M_i=0$ or $M_i=1$.

Let $\epsilon>0$ be a threshold for entering the game. A
decision rule for entering will be the following: at step $i$ enter
the game (buy $M_i=1$ of shares) if $\tilde p_i>\tilde S_{i-1}+\epsilon$;
pass the step otherwise (get $M_i=0$), where $\tilde S_{i-1}$ is randomized
past price. Thereby, we need changing checking rules depending on past outcomes~:
\[
I(p_i>S_{i-1}+\epsilon)=
  \left\{
    \begin{array}{l}
      1, \mbox{ if } p_i>S_{i-1}+\epsilon,
    \\
      0, \mbox{ otherwise. }
    \end{array}
  \right.
\]
It will follow from Theorem~\ref{univ-1b} (Section~\ref{gam-1a})
that there exists a randomized
algorithm computing forecasts calibrated almost surely in a modified sense~:
\begin{eqnarray*}\label{call-1bb}
\lim\limits_{n\to\infty}\frac{1}{n}\sum_{i=1}^{n}
I(\tilde p_i>\tilde S_{i-1}+\epsilon)(S_i-\tilde p_i)=0,
\end{eqnarray*}
where $\tilde p_i$ is a random forecast, $\tilde S_{i-1}$ is a
randomized past price of a stock, and $\epsilon>0$ is a
threshold for entering a game.

In Section~\ref{gam-0} we construct trading strategies based on
calibrated forecasts.

\section{Computing calibrated forecasts}\label{gam-1a}

Let $y_1,y_2,\dots$ be an infinite sequence of real numbers.
An infinite sequence of random variables $\tilde y_1,\tilde
y_2,\dots$ is called {\it a randomization} of $y_1,y_2,\dots$
if $E_n(\tilde y_n)=y_n$ for all $n$, where $E_n$ is the symbol of
mathematical expectation.

We specify a side information -- we add to the protocol of the game
signals $\bar x_1,\bar x_2,\dots$ given
online: for any $n$, a $k$-dimensional vector
$\bar x_n\in [0,1]^k$ is given to {\it Forecaster}
before he announces a forecast $\tilde p_n$.
We consider checking rules of general type:
\[
I(p,\bar x)=
  \left\{
    \begin{array}{l}
      1, \mbox{ если } (p,\bar x)\in {\cal S},
    \\
      0, \mbox{ otherwise, }
    \end{array}
  \right.
\]
where ${\cal S}\subseteq [0,1]^{k+1}$ and
$\bar x\in [0,1]^k$ is a signal. In Section~\ref{gam-0} we use
a set ${\cal S}=\{(p,x):p>x+\epsilon\}$,
where $p,x\in [0,1]$ and $\epsilon>0$. At any step $i$ we check
$(\tilde p_i,\tilde x_i)\in {\cal S}$, where $x_i=S_{i-1}$
and $\tilde p_i$, $\tilde x_i$ are randomization of $p_i$, $x_i$.

The following theorem on calibration is the main tool for technical
analysis presented in Section~\ref{gam-0}.
\begin{theorem}\label{univ-1b} Given $k$ an algorithm $f$ for computing
forecasts and a method of randomization can be constructed such that
for any sequence of real numbers $S_1,S_2,\dots$ and for any sequence
of $k$-dimensional signals $\bar x_1,\bar x_2,\dots$ the event
\begin{eqnarray}
\lim\limits_{n\to\infty}\frac{1}{n}\sum_{i=1}^{n}
I(\tilde p_i,\tilde x_i)(S_i-\tilde p_i)=0,
\label{call-1b}
\end{eqnarray}
has $Pr$-probability 1, where $Pr$ is a probability distribution
generated by a sequence of tuples $(\tilde p_i,\tilde x_i)$ of random
variables, $i=1,2,\dots$, and $I$ is the characteristic function of an
arbitrary subset ${\cal S}\subseteq [0,1]^{k+1}$. Here $\tilde p_i$
is the randomization of a forecast $p_i$ computed by the forecasting algorithm
$f$ and $\tilde x_i$ is obtained by independent randomization of each
coordinate $x_{i,j}$ of the vector $\bar x_i$, $j=1,\dots k$.
Also $\Var_n(\tilde p_n)\to 0$ and $\Var_n(\tilde x_{i,j})\to 0$
as to $n\to\infty$ for all $i$ and $j$.
\footnote
{
$\Var_n(\tilde p_n)=E_n(\tilde p_n-p_n)^2$.
}
\end{theorem}
{\it Proof.} We modify a weak calibration algorithm of
Kakade and Foster~\cite{KaF2004} using also ideas from Vovk~\cite{Vov2007}.

At first, we construct an $\Delta$-calibrated
forecasting algorithm, and after that we apply some double trick argument
for it.
\begin{lemma}\label{univ-1c}
Given $k$ an algorithm for computing forecasts and a method of randomization
can be constructed such that for any sequence of real numbers $S_1,S_2,\dots$
and for any sequence of signals $\bar x_1,\bar x_2,\dots$ the event
\begin{eqnarray*}
\limsup\limits_{n\to\infty}\frac{1}{n}\sum_{i=1}^{n}
I(\tilde p_i,\tilde x_i)(S_i-\tilde p_i)\le\Delta
\end{eqnarray*}
has $Pr$-probability 1, where $Pr$ and $I$ as in Theorem~\ref{univ-1b}.
Also $\Var_n(\tilde p_n)\le\Delta$ and $\Var_n(\tilde x_{i,j})\le\Delta$
for all $n$, for all $i$ and $j$.
\end{lemma}
{\it Proof}. At first we define a deterministic forecast and after that
we randomize it.

Divide the interval $[0,1]$ on subintervals of length $\Delta=1/K$
with rational endpoints $v_i=i\Delta$, where $i=0,1,\dots , K$.
Let $V$ denotes the set of these points.

Any number $p\in [0,1]$ can be represented as a linear
combination of two neighboring endpoints of $V$ defining
subinterval containing $p$~~:
$
p=\sum\limits_{v\in V}w_v(p)v=
w_{v_{i-1}}(p)v_{i-1}+w_{v_i}(p)v_i,
$
where $p\in [v_{i-1},v_i]$, $i=\lfloor p^1/\Delta+1\rfloor$,
$w_{v_{i-1}}(p)=1-(p-v_{i-1})/\Delta$, and
$w_{v_i}(p)=1-(v_i-p)/\Delta$.
Define $w_v(p)=0$ for all other $v\in V$.

In that follows we round some deterministic forecast $p_n$ to
$v_{i-1}$ with probability $w_{v_{i-1}}(p_n)$ and to $v_i$
with probability $w_{v_i}(p_n)$.
We also round the each coordinate $x_{n,s}$, $s=1,\dots k$, of a signal
$\bar x_n$ to
$v_{j_s-1}$ with probability $w_{v_{j_s-1}}(x_{n,s})$ and to $v_{j_s}$
with probability $w_{v_{j_s}}(x_{n,s})$,
where $x_{n,s}\in [v_{j_s-1},v_{j_s}]$.

Let also $W_v(Q_n)=w_{v^1}(p_n)w_{v^2}(\bar x_n)$,
where $v=(v^1,v^2)$, $v^1\in V$, $v^2=(v^2_1,\dots v^2_k)\in V^k$,
$w_{v^2}(\bar x_n)=\prod_{s=1}^k w_{v^2_s}(x_{n,s})$,
and $Q_n=(p_n,\bar x_n)$. For any $Q_n$, $W_v(Q_n)$ is a
probability distribution in $V^{k+1}$~:
$\sum\limits_{v\in V^{k+1}}W_v(Q_n)=1$.

In that follows we define a deterministic forecast $p_n$.
Let the forecasts $p_1,\dots , p_{n-1}$ already defined (put $p_1=1/2$).
Let us define for $v=(v^1,v^2)$ and $Q_i=(p_i,\bar x_i)$
$$
\mu_{n-1}(v)=\sum\limits_{i=1}^{n-1} W_v(Q_i)(S_i-p_i).
$$
We have
\begin{eqnarray}
(\mu_n(v))^2=(\mu_{n-1}(v))^2+
\nonumber
\\
+2W_v(Q_n)\mu_{n-1}(v)(S_n-p_n)+
(W_v(Q_n))^2(S_n-p^1_n)^2.
\label{m-1}
\end{eqnarray}
Summing (\ref{m-1}) by $v\in V^{k+1}$, we obtain:
\begin{eqnarray}
\sum\limits_{v\in V^{k+1}}(\mu_n(v))^2=\sum\limits_{v\in V^{k+1}}(\mu_{n-1}(v))^2+
\nonumber
\\
+2(S_n-p_n)\sum\limits_{v\in V^{k+1}}W_v(Q_n)\mu_{n-1}(v)+
\sum\limits_{v\in V^{k+1}}(W_v(Q_n))^2(S_n-p_n)^2.
\label{m21}
\end{eqnarray}
Change the order of summation:
\begin{eqnarray*}
\sum\limits_{v\in V^{k+1}}W_{v}(Q_n)\mu_{n-1}(v)=
\sum\limits_{v\in V^{k+1}}W_{v}(Q_n)
\sum\limits_{i=1}^{n-1} W_{v}(Q_i)(S_i-p_i)=
\nonumber
\\
=\sum\limits_{i=1}^{n-1}(\sum\limits_{v\in V^{k+1}}
W_{v}(Q_n)W_{v}(Q_i))(S_i-p_i)=
\nonumber
\\
=\sum\limits_{i=1}^{n-1}(\bar W(Q_n)\cdot\bar W(Q_i))(S_i-p_i)=
\sum\limits_{i=1}^{n-1}K(Q_n,Q_i)(S_i-p_i),
\end{eqnarray*}
where $\bar W(Q_n)=(W_v(Q_n):v\in V^{k+1})$,
$\bar W(Q_n)=(W_v(Q_n):v\in V^{k+1})$
be vectors of probabilities of rounding.
The dot product of corresponding vectors defines the kernel
\begin{eqnarray}
K(Q_n,Q_i)=K(p_n,\bar x_n,p_i,\bar x_i)=(\bar W(Q_n)\cdot\bar W(Q_i)).
\label{ker-1}
\end{eqnarray}

Let $p_n$ be equal to the root of the equation
\begin{eqnarray}
S_n(p_n)=\sum\limits_{v\in V}W_{v}(p_n,\bar x_n)\mu_{n-1}(v)=
\sum\limits_{i=1}^{n-1}K(p_n,\bar x_n,p_i,\bar x_i)(S_i-p_i)=0,
\label{equ-1}
\end{eqnarray}
if a solution exists. Otherwise, if the left hand-side of the equation
(\ref{equ-1}) (which is a continuous by $p_n$ function)
strictly positive for all $p_n$ define $p_n=1$,
define $p_n=0$ if it is strictly negative.
Announce $p_n$ as a deterministic forecast.

The third term of (\ref{m21}) is upper bounded by $1$. Indeed,
since $|S_i-p_i|\le 1$ for all $i$,
\begin{eqnarray*}
\sum\limits_{v\in V^{k+1}}(W_v(Q_n))^2(S_i-p_n)^2\le
\sum\limits_{v\in V^{k+1}}W_v(Q_n)=1.
\end{eqnarray*}
Then by (\ref{m21}),
$
\sum\limits_{v\in V^{k+1}}(\mu_n(v))^2\le n.
$
Recall that for any  $v\in V^{k+1}$
\begin{eqnarray}
\mu_n(v)=\sum\limits_{i=1}^n W_{v}(Q_i)(S_i-p_i).
\label{mmm-1}
\end{eqnarray}
Insert the term $I(v)$ in the sum (\ref{mmm-1}), where $I$ is
the characteristic function of an arbitrary set
${\cal S}\subseteq [0,1]^{k+1}$,
sum by $v\in V^{k+1}$, and exchange the order of
summation. Using Cauchy--Schwartz inequality for vectors
$(I(v):v\in V^{k+1})$, $(\mu_n(v):v\in V^{k+1})$ and
Euclidian norm, we obtain
\begin{eqnarray}\label{inn-2a}
\left|\sum\limits_{i=1}^n\sum\limits_{v\in V^{k+1}}W_{v}(Q_i)
I(v)(S_i-p_i)\right|=
\nonumber
\\
=\left|\sum\limits_{v\in V^{k+1}}I(v)
\sum\limits_{i=1}^n W_{v}(Q_i)(S_i-p_i)\right|\le
\nonumber
\\
\le\sqrt{\sum\limits_{v\in V^{k+1}}I(v)}
\sqrt{\sum\limits_{v\in V^{k+1}}(\mu_n(v))^2}\le\sqrt{|V^{k+1}|n}
\end{eqnarray}
for all $n$, where $|V^{k+1}|=1/\Delta^{k+1}$ -- is the cardinality of
the partition.

Let $\tilde p_i$ be a random variable taking values
$v\in V$ with probabilities $w_{v}(p_i)$ (only two of them are nonzero).
Recall that $\tilde x_i$ is a random variable taking
values $v\in V^k$ with probabilities $w_{v}(\bar x_i)$.

Let ${\cal S}\subseteq [0,1]^k$ and $I$ be its indicator function.
For any $i$ the mathematical expectation of a random variable
$I(\tilde p_i,\tilde x_i)(S_i-\tilde p_i)$ is equal to
\begin{eqnarray}\label{expe-1}
E(I(\tilde p_{i},\tilde x_i)(S_i-\tilde p_i))=
\sum\limits_{v\in V^{k+1}}W_{v}(Q_i)I(v)(S_i-v^1),
\end{eqnarray}
where $v=(v^1,v^2)$.

By the strong law of large numbers, for some $\mu_n=o(n)$
(as $n\to\infty$), $Pr$-probability of the event
\begin{eqnarray}
\left|\sum\limits_{i=1}^n I(\tilde p_{i},\tilde x_i)(S_i-\tilde p_i)-
\sum\limits_{i=1}^n E(I(\tilde p_{i},\tilde x_i)
(S_i-\tilde p_i))\right|\le\mu_{n}
\label{iii-1}
\end{eqnarray}
tends to 1 as $n\to\infty$. A form of of $\mu_{n}$ will be specified later.

By definition of deterministic forecast
\begin{eqnarray*}
\left|\sum\limits_{v\in V^{k+1}}W_{v}(Q_i)I(v)(S_i-p_i)-
\sum\limits_{v\in V^{k+1}}W_{v}(Q_i)I(v)(S_i-v^1)\right|<
\Delta
\end{eqnarray*}
for all $i$, where $v=(v^1,v^2)$.
Summing (\ref{expe-1}) by $i=1,\dots ,n$ and using the inequality
(\ref{inn-2a}), we obtain
\begin{eqnarray}
\left|\sum\limits_{i=1}^n
E(I(\tilde p_{i},\tilde x_i)(S_i-\tilde p_i))\right|=
\nonumber
\\
=\left|\sum\limits_{i=1}^n\sum\limits_{v\in V^{k+1}}
W_v(Q_i)I(v)(S_i-v^1)\right|
<\Delta n+\sqrt{|V^{k+1}|n}
\label{iinn-2}
\end{eqnarray}
for all $n$, where $|V^{k+1}|=1/\Delta^{k+1}$ is the cardinality
of the partition.

By (\ref{iinn-2}) and (\ref{iii-1}) we obtain that $Pr$-probability
of the event
\begin{eqnarray}
\left|\sum\limits_{i=1}^n
I(\tilde p_{i}, \tilde x_i)(S_i-\tilde p_i)\right|\le
\Delta n+\mu_n+\sqrt{n/\Delta^{k+1}}
\label{con-1}
\end{eqnarray}
tends to 1 as $n\to\infty$. In particular, $Pr$-probability of the event
\begin{eqnarray*}\label{cali-1}
\limsup\limits_{n\to\infty}\left|\frac{1}{n}\sum\limits_{i=1}^n
I(\tilde p_{i},\tilde x_i)(S_i-\tilde p_i)\right|\le\Delta
\end{eqnarray*}
is equal to 1. Lemma is proved.

To prove that (\ref{call-1b}) holds almost surely choose
a monotonic sequence of rational numbers
$
\Delta_1>\Delta_2>\dots
$
such that $\Delta_s\to 0$ as $s\to\infty$.
We also define an increasing sequence of natural numbers
$
n_{1}<n_{2}<\dots
$
For any $s$, we use on steps $n_{s}\le n<n_{s+1}$ the partition of $[0,1]$
on subintervals of length $\Delta_{s}$.

We choose $n_s$ such that
$n_s\ge\left(\frac{k+2}{2}\right)^2\Delta_s^{-(k+3)}$ for all $s$.
\footnote
{
This is the minimum point of (\ref{iinn-2}).
}
Put $n_0=0$ and $\Delta_0=1$.
Also, define the numbers $n_1,n_2,\dots$ such that the inequality
\begin{eqnarray}
\left|\sum\limits_{i=1}^{n} E(I(\tilde p_i,\tilde x_i)
(S_i-\tilde p_i))\right|\le 4(s+1)\Delta_{s} n
\label{exp1-1}
\end{eqnarray}
holds for all $n_s\le n\le n_{s+1}$ and for all $s$.

We define this sequence by mathematical induction on $s$.
Suppose that $n_s$ ($s\ge 1$) is defined such that the inequality
\begin{eqnarray}
\left|\sum\limits_{i=1}^{n} E(I(\tilde p_i,\tilde x_i)
(S_i-\tilde p_i))\right|\le 4s\Delta_{s-1}n
\label{exp1-1g}
\end{eqnarray}
holds for all $n_{s-1}\le n\le n_s$, and the inequality
\begin{eqnarray}
\left|\sum\limits_{i=1}^{n_s} E(I(\tilde p_i,\tilde x_i)
(S_i-\tilde p_i))\right|\le 4s\Delta_{s} n_s
\label{exp1-gg1}
\end{eqnarray}
also holds.
Let us define $n_{s+1}$. Consider all forecasts $\tilde p_i$
defined by the algorithm given above for discretization
$\Delta=\Delta_{s+1}$. We do not use first $n_s$ of these forecasts
(more correctly we will use them only in bounds (\ref{exp1-1f}) and
(\ref{exp1-1fg}); denote these forecasts ${\bf \hat p_1,\dots ,\hat p_{n_s}}$).
We add the forecasts $\tilde p_i$ for $i>n_s$ to the forecasts defined
before this step of induction (for $n_s$).
Let $n_{s+1}$ be such that the inequality
\begin{eqnarray}
\left|\sum\limits_{i=1}^{n_{s+1}} E(I(\tilde p_i,\tilde x_i)
(S_i-\tilde p_i))\right|\le
\left|\sum\limits_{i=1}^{n_s} E(I(\tilde p_i,\tilde x_i)
(S_i-\tilde p_i))\right|+
\nonumber
\\
+\left|\sum\limits_{i=n_s+1}^{n_{s+1}} E(I(\tilde p_i,\tilde x_i)
(S_i-\tilde p_i))+
\sum\limits_{i=1}^{n_s} E(I({\bf \hat p_i},\tilde x_i)
(S_i-{\bf \hat p_i}))\right|+
\nonumber
\\
+\left|\sum\limits_{i=1}^{n_s} E(I({\bf\hat p_i},\tilde x_i)
(S_i-{\bf \hat p_i}))\right|
\le 4(s+1)\Delta_{s+1}n_{s+1}~~~~~
\label{exp1-1f}
\end{eqnarray}
holds. Here the first sum of the right-hand side of the
inequality (\ref{exp1-1f}) is bounded by $4s\Delta_{s}n_{s}$ --
by the induction hypothesis (\ref{exp1-gg1}). The second and
third sums are bounded by $2\Delta_{s+1}n_{s+1}$ and by
$2\Delta_{s+1}n_s$, respectively. This follows from
(\ref{iinn-2}) and by choice of $n_s$.
The induction hypothesis (\ref{exp1-gg1}) is valid for
$$
n_{s+1}\ge\frac{2s\Delta_s+\Delta_{s+1}}{\Delta_{s+1}(2s+1)}n_s.
$$
Analogously,
\begin{eqnarray}
\left|\sum\limits_{i=1}^{n} E(I(\tilde p_i,\tilde x_i)
(S_i-\tilde p_i))\right|\le
\left|\sum\limits_{i=1}^{n_s} E(I(\tilde p_i,\tilde x_i)
(S_i-\tilde p_i))\right|+
\nonumber
\\
+\left|\sum\limits_{i=n_s+1}^{n} E(I(\tilde p_i,\tilde x_i)
(S_i-\tilde p_i))+
\sum\limits_{i=1}^{n_s} E(I({\bf \hat p_i},\tilde x_i)
(S_i-{\bf \hat p_i}))\right|+
\nonumber
\\
+\left|\sum\limits_{i=1}^{n_s} E(I(\hat p_i,\tilde x_i)
(S_i-{\bf\hat p_i}))\right|
\le 4(s+1)\Delta_sn~~~~~
\label{exp1-1fg}
\end{eqnarray}
for $n_s<n\le n_{s+1}$. Here the first sum of the right-hand
inequality (\ref{exp1-1f}) is also bounded by
$4s\Delta_{s}n_{s}\le 4s\Delta_{s}n$ -- by the induction
hypothesis (\ref{exp1-gg1}). The second and the third sums are
bounded by $2\Delta_{s+1}n\le 2\Delta_sn$ and by
$2\Delta_{s+1}n_s\le 2\Delta_sn$, respectively. This follows
from (\ref{iinn-2}) and from choice of $\Delta_s$. The
induction hypothesis (\ref{exp1-1g}) is valid.

By (\ref{exp1-1}) for any $s$
\begin{eqnarray}
\left|\sum\limits_{i=1}^{n} E(I(\tilde p_i,\tilde x_i)
(S_i-\tilde p_i))\right|\le 4(s+1)\Delta_sn
\label{exp1-1er}
\end{eqnarray}
for all $n\ge n_s$ if $\Delta_{s}$ satisfies
the condition $\Delta_{s+1}\le\Delta_{s}(1-\frac{1}{s+2})$ for all $s$.

By the law of large numbers (\ref{sem-h-1ddd}), the relation
(\ref{iii-1}) can be specified:
\begin{eqnarray}
Pr\left\{\sup\limits_{n\ge n_s}\left|\frac{1}{n}
\sum\limits_{i=1}^n V_i
\right|>\Delta_s\right\}\le(\Delta_s)^{-2} e^{-2n_s\Delta_s^2}
\label{iii-1ss}
\end{eqnarray}
for all $s$, where
$
V_i=I(\tilde p_i,\tilde x_i)(S_i-\tilde p_i)-
E(I(\tilde p_i,\tilde x_i)(S_i-\tilde p_i))
$
is a sequence of martingale--differences.

Combining (\ref{exp1-1er}) with (\ref{iii-1ss}), we obtain
\begin{eqnarray}
Pr\left\{\sup\limits_{n\ge n_{s}}\left|\frac{1}{n}\sum\limits_{i=1}^n
I(\tilde p_i,\tilde x_i)(S_i-\tilde p_i)\right|\ge
(4s+5)\Delta_{s}\right\}\le
(\Delta_s)^{-2}e^{-2n_{s}\Delta_{s}^{2}}
\label{cali-1rr}
\end{eqnarray}
for all $s$.
The series $\sum_{s=1}^\infty
(\Delta_s)^{-2}e^{-2n_{s}\Delta_{s}^{2}}$ is convergent
if $n_s$ satisfies
$$
n_s\ge\frac{\ln s+2\ln\ln s-2\ln(\Delta_s)}{2\Delta_s^2}
$$
for all $s$. Let also $\Delta_{s}=o(1/s)$ as $s\to\infty$.
Then Borel--Cantelli Lemma implies convergence of
(\ref{call-1b}) almost surely.

It is easy to verify that the sequences $n_{s}$ and $\Delta_{s}$ satisfying
all the conditions above exist.

\section{Applications to technical analysis}\label{gam-0}

\subsection{Simple trading for a rise}\label{gam-1}

Let $S_1, S_2,\dots $ be a sequence of a stock prices.
At any step $i$ we use one-dimensional signals $x_i=S_{i-1}$, $i=1,2\dots$,
and an indicator function $I(p_i>x_i+\epsilon)$, where $\epsilon$
is a parameter.
\footnote
{
In other approach we can consider a sequence of signals
$\epsilon_1,\epsilon_2,\dots$.
}

At the end of the trading period {\it Speculator} receives a gain or suffer
loss $\Delta S_I=S_i-S_{i-1}$ for one share of the stock.
The total gain or loss is equal to
$$
{\cal K}_n=\sum\limits_{i=1}^n
I(\tilde p_i>\tilde S_{i-1}+\epsilon)\Delta S_i,
$$
where $\tilde p_i$ and $\tilde S_{i-1}$ are randomized forecast and
past price of the stock respectively.

Let us specify details of rounding.
The expression $\Delta n+\sqrt{n/\Delta^{k+1}}$
from (\ref{con-1}) takes its minimal value for
$\Delta=(\frac{k+1}{2})^\frac{2}{k+3}n^{-\frac{1}{k+3}}$.
In this case, the right-hand
side of the inequality (\ref{iinn-2}) is equal to
$\Delta n+\sqrt{n/\Delta^{k+1}}=2\Delta n=
2(\frac{k+1}{2})^\frac{2}{k+3}n^{1-\frac{1}{k+3}}$.

We have $k=1$, and hence, we use at any step $n$ the rounding
$\Delta_s=n_s^{-1/4}$, where $s$ is such that $n_s<n\le n_{s+1}$.

We write $A\sim B$ if positive constants $c_1$ and $c_2$ exist such that
$c_1B\le A\le c_2B$ for all values of parameters from the expressions $A$ and $B$.

Define $n_s=s^M$ and $\Delta_s=s^{-M/4}$, where
$M$ is a positive integer number. Then $s\sim n_s^{1/M}$ (the constants
$c_1$ and $c_2$ depend on $M$).

Easy to verify that all requirements for $n_s$ and $\Delta_s$
given in Section~\ref{gam-1a} are valid.

By (\ref{cali-1rr}) we can define $\mu_{n}=(4s+5)\Delta_{s}n$,
where $s$ is such that $n_{s}<n\le n_{s+1}$.
For $n_s<n\le n_{s+1}$ it holds $n\sim n_s$, hence,
$\mu_n\sim n^{3/4+1/M}$.

We represent the total gain by $n$ steps in a form
\begin{eqnarray}
{\cal K}_n=\sum\limits_{\tilde p_i>\tilde S_{i-1}+\epsilon}\Delta S_i=
\sum\limits_{i=1}^n I(\tilde p_i>\tilde S_{i-1}+\epsilon)
(S_i-\tilde p_i)+
\nonumber
\\
+\sum\limits_{i=1}^n I(\tilde p_i>\tilde S_{i-1}+\epsilon)
(\tilde S_{i-1}-S_{i-1})+
\nonumber
\\
+\sum\limits_{i=1}^nI(\tilde p_i>\tilde S_{i-1}+\epsilon)
(\tilde p_i-\tilde S_{i-1}).
\label{av-1a}
\end{eqnarray}
By (\ref{con-1}) the probability that the first addend of the
sum (\ref{av-1a}) is more than
$
-(\Delta_s n+\mu_n+2\sqrt{n/\Delta_s^2})
$
tends to 1 as $n\to\infty$, where $s$ is such that $n_s<n\le n_{s+1}$.

According to Section~\ref{marti-1} the probability that the second addend
of the sum (\ref{av-1a}) is more than $-\Delta_s n$ tends to 1 as $n\to\infty$.
By definition the third addend of the sum (\ref{av-1a})
is more than
$\epsilon\sum\limits_{i=1}^n I(\tilde p_i>\tilde S_{i-1}+\epsilon)$
for all $n$.


Then the probability that that the average income per one gamble $k_n$ satisfies
\begin{eqnarray}
k_n=\frac{{\cal K}_n}{\sum\limits_{i=1}^n
I(\tilde p_i>\tilde S_{i-1}+\epsilon)}\ge
\left(\epsilon -
\frac{2\Delta_s n+\mu_{n}+2\sqrt{n/\Delta_s^2}}
{\sum\limits_{i=1}^n I(\tilde p_i>\tilde S_{i-1}+\epsilon)}\right)\sim
\nonumber
\\
\sim
\left(\epsilon -
\frac{3n^{3/4}+n^{3/4+1/M}}
{\sum\limits_{i=1}^n I(\tilde p_i>\tilde S_{i-1}+\epsilon)}\right)
\label{av-2}
\end{eqnarray}
tends to 1 as $n\to\infty$.
Using inequalities of Section~\ref{marti-1}
and definition of $\Delta_s$ and $\mu_n$, one can check that
the corresponding convergence rate is $e^{-c\sqrt{n}}$, where $c>0$.


We summarize this result in the following proposition.
\begin{proposition}
A randomized trading strategy exists such that
given $0<\epsilon<1$ and  $0<\gamma<1$ with (internal) probability
$1-e^{-c'\sqrt{n}}$ if
\begin{eqnarray}
\sum\limits_{i=1}^n I(\tilde p_i>\tilde S_{i-1}+\epsilon)\ge
\frac{cn^{3/4+\nu}}{\gamma\epsilon},
\label{av-4}
\end{eqnarray}
then $k_n\ge (1-\gamma)\epsilon$,
where $\nu=1/M$, $c'$ and $c$ are positive constants.
\end{proposition}


\subsection{Trading with a limited risk}\label{gam-2}

The most important requirement for a trading strategy is guarantee
conditions. We present a defensive strategy for {\it Speculator} in sense
of Shafer and Vovk's book~\cite{ShV2001}.
This means that starting with some initial
capital {\it Speculator} never goes to debt and receives a gain
when a sufficiently long subsequence of forecasts like (\ref{av-4}) exists.

We modify the strategy given in Section~\ref{gam-1} to a defensive
strategy.

Let ${\cal K}_0>0$ be a starting capital of {\it Speculator}.
Define $M_i=\delta {\cal K}_{i-1}$, where
${\cal K}_{i-1}$ is the capital of {\it Speculator} at step $i-1$ and $\delta$
is a parameter such that $0\le\delta\le 1$.
As usual, we suppose that all prices a scaled such that
$0\le S_{i-1}\le 1$ и $S_1=S_0$.

{\it Speculator's} capital after $i$th step is equal to
\begin{eqnarray}\label{capp-1}
{\cal K}_{i}={\cal K}_{i-1}+\delta {\cal K}_{i-1}\Delta S_i.
\end{eqnarray}
At any step $n$ the logarithm of the capital is equal to
\begin{eqnarray}
\ln {\cal K}_n=\ln {\cal K}_0+
\sum\limits_{i=1}^{n}I(\tilde p_i>\tilde S_{i-1}+\epsilon)
\ln (1+\delta \Delta S_i)\ge
\nonumber
\\
\ge
\ln{\cal K}_0+\delta\sum\limits_{i=1}^{n}I(\tilde p_i>\tilde S_{i-1}+\epsilon)
\Delta S_i-
\nonumber
\\
-
\delta^2\sum\limits_{i=1}^{n}
I(\tilde p_i>\tilde S_{i-1}+\epsilon)(\Delta S_i)^2.
\label{iinc-2a}
\end{eqnarray}
Here we have used the inequality $\ln(1+x)\ge x-x^2$ for $|x|\le 1$.

Let $\tilde L_n=\sum\limits_{i=1}^{n}I(\tilde p_i>\tilde  S_{i-1}+\epsilon).$
Since $|\Delta S_{i-1}|\le 1$, a bound
\begin{eqnarray}
\sum\limits_{i=1}^{n}I(p_i>\tilde S_{i-1}+\epsilon)(\Delta S_i)^2
\le\tilde L_n
\label{bou-1}
\end{eqnarray}
is valid for all $n$.

By (\ref{av-2}) probability of the event
\begin{eqnarray}
\sum\limits_{i=1}^{n}I(p_i>\tilde S_{i-1}+\epsilon)\Delta S_i\ge
\epsilon\tilde L_n-cn^{3/4+1/M}
\label{iinc-2aa}
\end{eqnarray}
tends to 1 as $n\to\infty$, where $c$ is a positive constant.

Denote
\begin{eqnarray*}
{\rm \overline{Var}}_n(S)=
\frac{1}{\tilde L_n}
\sum\limits_{i=1}^{n}I(\tilde p_i>\tilde S_{i-1}+\epsilon)(\Delta S_i)^2.
\end{eqnarray*}

For practical applications, where ${\rm \overline{Var}}_n(S)\ll 1$,
we can replace in (\ref{bou-1}) $\tilde L_n$
on $\tilde L_n{\rm \overline{Var}}_n(S)$.
Then by (\ref{iinc-2a}) and (\ref{iinc-2aa}) probability of the event
\begin{eqnarray}
\ln {\cal K}_n\ge\ln{\cal K}_0+
\delta(\epsilon\tilde L_n-
cn^{3/4+1/M})-
\delta^2 {\rm \overline{Var}}_n(S)\tilde L_n
\label{iinc-2}
\end{eqnarray}
tends to 1 as $n\to\infty$.
Therefore, probability of the event
\begin{eqnarray}
{\cal K}_n\ge
{\cal K}_0 \exp \left(\delta\left(\tilde L_n
(\epsilon-\delta{\rm \overline{Var}}_n(S))-cn^{3/4+1/M}\right)\right)
\label{inc-1}
\end{eqnarray}
tends to 1 as $n\to\infty$.

We summarize the result of this section in the following proposition.
\begin{proposition}
A randomized trading strategy exists such that
given $0<\epsilon<1$ and $0<\delta<1$ with (internal) probability
$1-e^{-c'\sqrt{n}}$ the capital ${\cal K}_n$ of {\it Speculator}
has a lower bound (\ref{inc-1}), where $c'$ and $c$ are positive constants.
The capital increases if
\begin{eqnarray*}
\sum\limits_{i=1}^{n}I(\tilde p_i>\tilde  S_{i-1}+\epsilon)\ge
\frac{cn^{3/4+\nu}}{(\epsilon-\delta {\rm \overline{Var}}_n(S))},
\end{eqnarray*}
where $\nu=1/M$. Also, ${\cal K}_n>0$ for all $n$.
\end{proposition}

\section{Conclusion}

Calibration is an intensively developing area of recent research where several
algorithms for computing calibrated forecasts were developed. It is
attractive to find some practical applications of these results.
Using calibrated forecasts for constructing short-term trading strategies
in Stock Market looks very natural.

In this paper, we construct such strategies and
perform numerical experiments. These experiments show
a positive return for six main Russian stocks, and for two stocks we
receive a gain even when transaction costs are subtracted.

To construct trading strategies we develop a more general notion of calibration
and prove convergence results for it. We also present some sufficient
conditions under which our trading strategies receive a gain.

\begin{figure}[t]
\centering\includegraphics[height=60mm,width=135mm,clip]{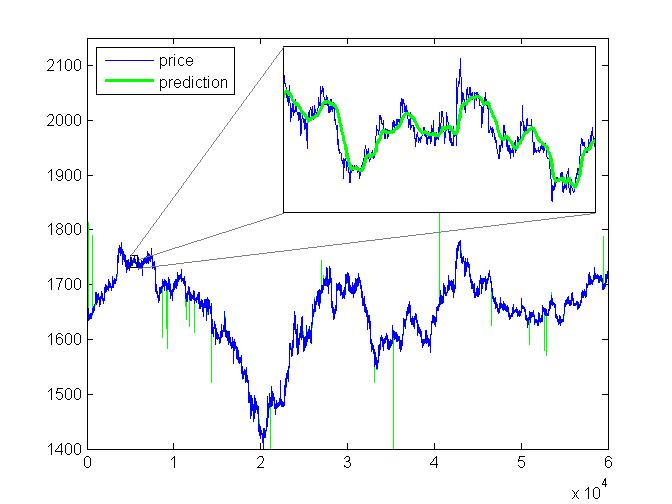}
{\small Fig.1 Calibrated forecasts of prices of LKOH stock}

                  \end{figure}

\begin{figure}[t]
\centering\includegraphics[height=60mm,width=135mm,clip]{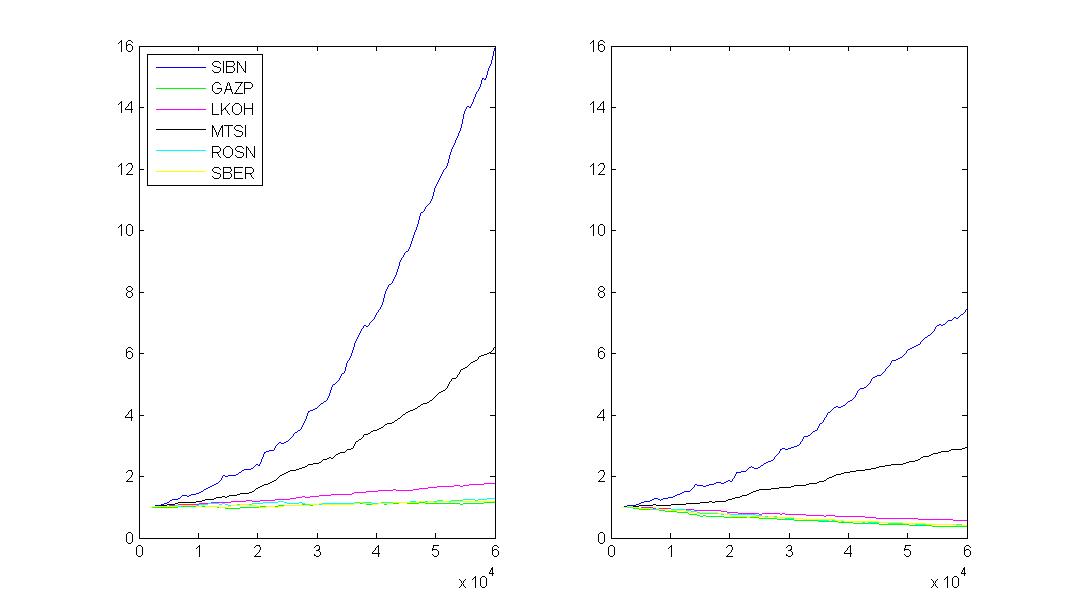}
{\small Fig.2 Capitals of speculators playing for a rise on six Russian stocks
(with no transaction costs -- on the left figure,
with transaction costs $0.01\%$ -- on the right figure)}
                  \end{figure}


\begin{figure}[t]
\centering\includegraphics[height=60mm,width=135mm,clip]{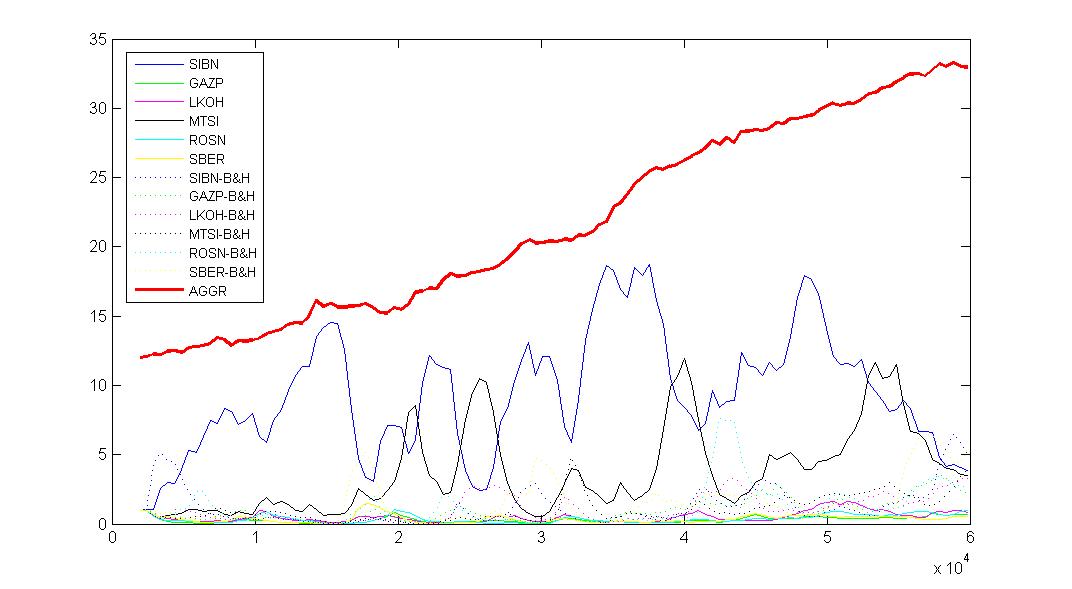}
{\small Fig.3 Capital of the weighted average of 12 strategies
(two strategies for each stock -- calibration strategy and ``buy and hold''
(B$\&$H) strategy}

\end{figure}



\appendix
\section{Appendix}
\subsection{Large deviation inequality for martingales}\label{marti-1}

A sequence $V_1,V_2,~\dots$ is called martingale-difference with respect
to a sequence of random variables $X_1,X_2,~\dots$ if for any $i>1$
the random variable $V_i$ is a function of $X_1,~\dots ,~X_i$ and
$
E(V_{i+1}|X_1,~\dots ,~X_i)=0
$
almost surely.
The following inequalities are consequences of Hoffding-Azuma
inequality~\cite{Ces2006}:

Let $V_1,V_2,~\dots$ be a martingale--difference with respect to
$X_1,X_2,~\dots$, and $V_i\in [A_i,A_i+1]$ for some random variable
$A_i$ measurable with respect to $X_1,~\dots ,~X_i$. Let
$S_n=\sum\limits_{i=1}^n V_i$. Then for any $t>0$
\begin{eqnarray}\label{sem-h-1dd}
P\left\{\left|\frac{S_n}{n}\right|>t\right\}\le 2e^{-2nt^2}
\end{eqnarray}
for all $n$. A strong law of large numbers is also holds: for any $t$
\begin{eqnarray}\label{sem-h-1ddd}
P\left\{\sup\limits_{k\ge n}\left|\frac{S_k}{k}\right|>t\right\}\le
t^{-2}e^{-2nt^2}
\end{eqnarray}
for all $n$.
Since the series of the exponents from the right-hand side of the inequality
(\ref{sem-h-1dd}) convergent, by Borel--Cantelli Lemma
we obtain the martingale strong law of large numbers
\begin{eqnarray*}\label{sem-h-1dddd}
P\left\{\lim\limits_{n\to\infty}\frac{S_n}{n}=0\right\}=1.
\end{eqnarray*}

\subsection{Numerical experiments}

In the numerical experiments, we have used historical data in form of
per minute time series of prices of six main
stocks of Russian Stock Market in 2010
(From 2010-03-26T10:31 to 2010-09-16T12:15).
downloaded from FINAM site: ${\it www.finam.ru}$.
Number of trading points in each game is $6\cdot 10^4$ min. In our experiments,
we dealing only for a rise starting with the same initial
capital ${\cal K}_0$.

We have used the threshold $\epsilon=\epsilon'\sigma$, where $\sigma$ is
the standard deviation of a price calculating using some sliding window,
$0<\epsilon'<1$. A kernel $K(p,p')=\cos(\pi(p-p'))$ was used as
a smooth approximation of (\ref{ker-1}).

Results of numerical experiments are shown in Table 1. In the first column,
ticker symbols of six stocks from Russian Stock Market are shown.
The second column contains the frequencies of steps $i$
where $p_i>S_{i-1}+\epsilon$.
In the third column, the average duration of a gamble is shown.
We sell all shares of a stock at step $i$ in case $\tilde p_i\le S_{i-1}+\epsilon$
or $S_i\le S_{i-1}$. In fourth and in fifth columns, a relative return
(in percentage wise on initial capital)
for six main stocks from Russian Stock Market in 2010. We have used
a transaction cost at the rate $0.01\%$. In the sixth column, a return
of ``buy and hold'' strategy is shown. By this strategy, we buy a
holdings of shares for ${\cal K}_0$ and sell them at the end of the trading
period.

On Fig.1 the evolution of LKOH prices and their predictions are shown.
On Fig. 2 the relative returns of calibration strategies for all
six stocks are shown (without and with transaction costs).
On Fig.3 a relative return of short-term trading for six stocks
are shown including calibration and buy and hold strategy for each stock.
The extra bold line represents a relative return
of some averaging strategy AGGR. This strategy is similar to
the Freund and Shapire~\cite{FrS97} exponential weighting
algorithm, where one-day steps are used (see also the last line of Table 1).

{\bf Table 1}.
Relative return in percentage wise on capital used in dealing for a rise
for six main stocks of Russian Stock Market in 2010 and for the
aggregating strategy AGGR.
\\
\\
\begin{tabular}{|l|l|l|l|l|l|l|l|}
\hline
Ticker&frequency&average&without&with&buy\\
symbol&of entry&duration&transaction&transaction&and\\
of a stock&points&of a gamble&costs&costs&hold\\
\hline
\hline
GAZP&0.100&1.88&$15.73\%$&$-63.67\%$&$-6.30\%$\\
\hline
LKOH&0.099&1.85&$78.87\%$&$-43.53\%$&$2.19\%$\\
\hline
MTSI&0.065&2.51&$527.05\%$&$196.15\%$&$1.15\%$\\
\hline
ROSN&0.097&1.86&$27.25\%$&$-58.66\%$&$-12.88\%$\\
\hline
SBER&0.092&1.94&$19.72\%$&$-58.86\%$&$-2.61\%$\\
\hline
SIBN&0.066&2.86&$1504.39\%$&$646.01\%$&$-21.94\%$\\
\hline
AGGR&&&$761.17\%$&$321.67\%$&\\
\hline
\end{tabular}

\end{document}